%% file: main.tex
\documentclass{article}
\usepackage[utf8]{inputenc}

\usepackage{amsmath,amsfonts,amssymb,amsthm}

\usepackage{graphicx}
\usepackage{epstopdf}
\AppendGraphicsExtensions{.tif}
\usepackage{mathtools, nccmath}

\usepackage{url}
\usepackage{xparse}
\DeclareUnicodeCharacter{2061}{}
\DeclareUnicodeCharacter{2212}{}
\DeclarePairedDelimiterX{\set}[1]{\{}{\}}{\setargs{#1}}
\NewDocumentCommand{\setargs}{>{\SplitArgument{1}{;}}m}
{\setargsaux#1}
\NewDocumentCommand{\setargsaux}{mm}
{\IfNoValueTF{#2}{#1} {#1\,\delimsize|\,\mathopen{}#2}}

\title{Epistemic Deep Learning}
\author{Shireen Kudukkil Manchingal \and Fabio Cuzzolin}

\begin{document}
\maketitle

\input{abstract}
\input{intro}
\input{sota}
\input{epilearning}

\input{randomsetcnn}

\input{loss}
\input{evaluation}

\input{experiments}

\input{conclusion}

\bibliographystyle{plain}
\bibliography{bibliography}

\end{document}

%% file: abstract.tex
\begin{abstract}
The belief function approach to uncertainty quantification as proposed in the Demspter-Shafer theory of evidence is established upon the general mathematical models for set-valued observations, called \emph{random sets}. Set-valued predictions are the most natural representations of uncertainty in machine learning. In this paper, we introduce a concept called \emph{epistemic deep learning} based on the random-set interpretation of belief functions to model epistemic learning in deep neural networks. We propose a novel random-set convolutional neural network for classification that produces scores for sets of classes by learning set-valued ground truth representations. We evaluate different formulations of entropy and distance measures for belief functions as viable loss functions for these random-set networks. We also discuss methods for evaluating the quality of epistemic predictions and the performance of epistemic random-set neural networks. We demonstrate through experiments that the epistemic approach produces better performance results when compared to traditional approaches of estimating uncertainty.
\end{abstract}

%% file: intro.tex
\section{Introduction}

Despite its recent rise in popularity, in its current form, AI cannot confidently make predictions robust enough to stand the test of data generated by processes different from those studied at training time, even by tiny details, as shown by `adversarial' results able to fool deep neural networks \cite{papernot2016limitations}. While recognising this issue under different names (e.g. \emph{overfitting} \cite{roelofs2019meta} or \emph{model adaptation} \cite{li2020model}), traditional machine learning seems unable to address it in non-incremental ways. As a result, AI systems suffer from brittle behaviour, and find it difficult to operate in new situations, e.g. adapting an autonomous vehicle to driving in heavy rain or coping with other road users' different styles of driving \cite{9493770}.

Among different sources of uncertainty in machine learning, epistemic uncertainty - the type of uncertainty observed when there is a lack of information, is fundamental in order to perform trustworthy model-based inference. There have been many advances towards quantifying epistemic uncertainty in deep learning models over the past years. The limitations of classical probability and both its frequentist and Bayesian interpretations have led to the development of a number of alternatives. Extensions of classical probability theory have been proposed, starting from de Finetti's pioneering work on subjective probability \cite{DeFinetti74}.
Other formalisms include possibility-fuzzy set theory \cite{Zadeh78,Dubois90}, probability intervals
\cite{halpern03book}, credal sets \cite{levi80book,kyburg87bayesian}, monotone capacities \cite{wang97choquet}, random sets \cite{Nguyen78} and imprecise probability theory \cite{walley91book}. New original foundations of subjective probability in behavioural terms \cite{walley00towards} or by means of game theory \cite{shafer01book} have been introduced.

In this paper, we focus in particular on \emph{random set} representations as a special case. As they assign probability values to sets of outcomes directly, they naturally model the fact that observations almost invariable come in the form of sets. In the case of finite target spaces (i.e., for classification) we will model uncertainty using \emph{belief functions} \cite{Shafer76}, the finite incarnation of random sets. Based on belief functions, the Dempster-Shafer theory of evidence \cite{Shafer76} provides a general framework for modeling epistemic uncertainty by combining evidence from multiple sources for parameters modeled by focal sets with associated degrees of belief. Dempster-Shafer theory is a generalization of Bayesian inference \cite{smets86bayes} in which epistemic uncertainty can be modeled by single probability distributions and assigning subjective priors to probabilities to incorporate expert knowledge.

\subsection{Contributions} \label{sec:contributions}

The main contributions of the paper are the following:
\begin{itemize}
    \item 
    The formulation of a novel epistemic deep learning framework, centred on the explicit modelling of the epistemic uncertainty induced by training data of limited size and quality.
    \item
    A baseline epistemic CNN architecture for multi-label image classification.
    \item
    The formulation of novel loss functions for the training of epistemic classification networks, obtained by generalising the notions of Kullback-Leibler divergence and entropy.
    \item
    The formulation of a general approach to assessing epistemic predictions, which includes Bayesian and traditional deep learning as special cases.
    
\end{itemize}

\subsection{Paper outline} \label{sec:outline}

The paper is structured as follows. In Sec. \ref{sec:sota} we review the way epistemic uncertainty is modelled in machine learning. Sec. \ref{sec:epistemic-dl} introduces the concept of epistemic deep learning, discusses how epistemic uncertainty can be modelled at either the target space level or the parameter space level. Sec. \ref{sec:rs-cnns}, in particular, proposes a CNN architecture for epistemic classification in the random set setting. Sec. \ref{sec:losses} derives and introduces various generalisations of the classical cross entropy loss to random sets, based on the KL divergence and entropy. Sec. \ref{sec:evaluation} introduces a general way of evaluating epistemic predictions and compares it with traditional evaluation. Sec. \ref{sec:experiments} demonstrates results on four datasets suitably adapted for this purpose of both our epistemic CNN and traditional baselines, which confirm the proven dominance results. Sec. \ref{sec:conclusions} concludes the paper.

%% file: sota.tex
\section{State of the art} 
\label{sec:sota}

Researchers have recently taken some tentative steps to model uncertainty in machine learning and artificial intelligence. Scientists have adopted measures of epistemic uncertainty \cite{kendall2017uncertainties} to refuse or delay making a decision \cite{geifman2017selective}, take actions specifically directed at reducing uncertainty (as in active learning \cite{aggarwal2014active}), or exclude highly-uncertain data at decision making time \cite{kendall2017uncertainties}. Deep networks that abstain from making predictions are one such paradigm \cite{geifman2017selective}. The use of Gaussian processes \cite{hullermeier2021aleatoric} have also been proposed. However, most of the sparse attempts made so far to incorporate uncertainty in neural predictions \cite{hullermeier2021aleatoric,geifman2017selective} have been made from a Bayesian stance, which only captures aleatory uncertainty, and do not fully model epistemic uncertainty \cite{cuzzolin2020geometry}. In this section, we categorize the state-of-the-art uncertainty quantification techniques into four methods, namely, set-wise learning, bayesian inference, evidential learning, and conformal prediction.

\subsection{Set-wise learning} \label{sec:sota-set}

As a form of set-wise learning, epistemic learning relates to other approaches which learn sets of models \cite{10.1145/1015330.1015432} (although differing in its rationale) and to minimax approaches to domain adaptation \cite{NIPS2012_46489c17}, which employ minimax optimisation to learn models adapted to data generated by any probability distribution within a “safe” family. 

Zaffalon proposed Naive Credal Classifiers(NCC) \cite{Zaffalon}\cite{Zaffalon2}, an extension of the naive Bayes classifier to credal sets, where imprecise data probabilities are included in models in the form of sets of classes. Antonucci \cite{ANTONUCCI2017320} presents graphical models as generalisation of the NCCs to multilabel data. Expected utility maximization algorithms such as finding the Bayes-optimal prediction \cite{Mortier} and classification with reject option for risk aversion \cite{ijcai2018-706} are also based on set-valued predictions.

Much work has been done in unsupervised learning, and specifically in clustering by Denoeux and co-authors \cite{masson}\cite{evclus}. The clustering of belief functions, in particular, has been extensively studied by Schubert \cite{Schubert}. In classification, evidence-theoretical classifiers have been developed, based both on Smets’s generalised Bayes theorem \cite{DBLP:journals/tsmc/KimS95} and on distance functions, e.g. evidential K-NNs\cite{knearest}. Random forest approaches based on imprecise probabilities have been brought forward by Abellan and co-authors \cite{Abelln2017ARF} but also by Utkin \cite{UTKIN2020112978}. Classification with partial data has been studied by various authors \cite{10.1007/11518655_80}\cite{TABASSIAN20121698}, while a large number of papers have been published on decision trees in the belief function framework \cite{elouedi}\cite{VANNOORENBERGHE2004179}. Significant work in the neural network area was conducted, once again, by Denoeux \cite{denoeux}, and by Fay et al. \cite{10.1007/11829898_18}. A credal bagging approach was proposed in \cite{Francois_resampleand}. Classification approaches based on rough sets were proposed by Trabelsi et al \cite{Trabelsi2011ClassificationSB}. Ensemble classification \cite{burger} is another area in which uncertainty theory has been quite impactful, as the problem is one of combining the results of different classifiers. Important contributions have been made by Xu et al.\cite{155943} and by Rogova \cite{ROGOVA1994777}. Regression, on the other hand, was considered by belief theorists only relatively recently \cite{Laanaya2010338}\cite{7885577}.

\subsection{Bayesian deep learning} \label{sec:sota-bayesian}

Bayesian approach in deep learning was pioneered by Neal\cite{Neal1995BayesianLF}, MacKay \cite{10.1162/neco.1992.4.3.448} and Buntine and Weigend \cite{Buntine1991BayesianB}. Some literature in these papers establishes link between Bayesian Neural Networks and Gaussian processes. The Bayesian method to estimating uncertainty\cite{10.1162/neco.1992.4.3.448} is a probabilistic approach that is well-established in machine learning. Bayesian networks apply posterior inference on traditional neural network architecture and provide estimates of uncertainty in predictions.
Since the posterior is complex to compute analytically, in practice, different methods that try to mimic the posterior using a simpler, tractable family of distributions are employed. Approximation of Bayesian deep learning models are done using techniques such as Monte Carlo Markov Chain (MCMC) \cite{NIPS1992_f29c21d4}\cite{4767596}, variational inference(VI) \cite{jordan1999introduction}\cite{NIPS2011_7eb3c8be}, dropout VI \cite{https://doi.org/10.48550/arxiv.1506.02158}, or using an ensemble of independently trained models, to infer expectation and variance over stochastic passes where large variance measures indicate the model’s epistemic uncertainty. Early variational approximations to Bayesian network weights were described by Hinton \cite{10.1145/168304.168306} and Barber \cite{Barber1998EnsembleLI}.

Bayesian learning has some limitations \cite{DBLP:journals/corr/abs-2007-06823}. The computational cost for sampling during inference is high, limiting the potential for use in real-time applications. Additionally, it is difficult to choose an appropriate model prior and approximate inference scheme to design a model posterior.

\subsection{Evidential approaches} \label{sec:sota-evidential}

Within a proper epistemic setting, a significant amount of work has been done by Denoeux and co-authors, and Liu et al. \cite{Liu2012291}, on unsupervised learning and clustering in particular in the belief function framework \cite{Masson20081384}. Quite a lot of work has been done on ensemble classification in the evidential framework \cite{155943} (in particular for neural networks \cite{ROGOVA1994777}), decision trees \cite{elouedi00decision}, K-nearest neighbour classifiers \cite{Denoeux2008classic}, and more recently on evidential deep learning classifiers able to quantify uncertainty \cite{tong2021evidential}. Tong et al.\cite{tong2021evidential} proposes a convolutional neural network based on Dempster-Shafer theory called the evidential deep-classifier employs utility functions from decision theory to assign utilities on mass functions derived from input features to produce set-valued observations. Another recent approach by Sensoy et al.\cite{sensoy} proposes an evidential deep learning classifier to estimate uncertainty in the Dirichlet representation. This work is based on subjective logic and learning to form subjective opinions by minimizing the Kullback-Leibler divergence over a uniform Dirichlet distribution.

\subsection{Conformal prediction} 
\label{sec:sota-conformal}

Conformal prediction \cite{vovk2005algorithmic}\cite{shafer2008tutorial} provides a framework for estimating uncertainty by applying a threshold on the error the model can make to produce prediction sets, irrespective of the underlying prediction problem. Some of the early work in conformal prediction were formed by Gammerman, Vovk and Vapnik \cite{10.5555/2074094.2074112} in the 1990s. Balasubramanian et al.\cite{balasubramanian2014conformal} presented a more recent work on conformal methods in machine learning. In machine learning, the conformal method involves computing a non-conformity measure to compare between test and training data and achieve better accuracy. Different variants of conformal predictors are described in papers by Saunders et al.\cite{Saunders1999TransductionWC}, Nouretdinov et al.\cite{Nouretdinov01ridgeregression}, Proedrou et al.\cite{10.1007/3-540-36755-1_32} and Papadopoulos et al.\cite{10.5555/1712759.1712773}. Since the computational inefficiency of conformal predictors posed a problem for their use in deep learning networks, Inductive Conformal Predictors(ICPs) were proposed by Papadopoulos et al.\cite{10.1007/3-540-36755-1_29}\cite{ICP}. Venn Predictors \cite{NIPS2003_10c66082}, cross-conformal predictors \cite{crossconformal} and Venn-Abers predictors \cite{https://doi.org/10.48550/arxiv.1211.0025} were introduced in distribution-free uncertainty quantification using the conformal approach.

Nevertheless, these attempts to introduce concepts from uncertainty theory into machine learning have so far achieved limited impact, possibly because of the lack of a solid underlying principle.

%% file: epilearning.tex
\section{Epistemic deep learning} 
\label{sec:epistemic-dl}

In this paper, we introduce an \emph{epistemic deep learning} concept based on random sets and belief functions. Our approach to developing an \emph{epistemic artificial intelligence} theory is based on learning from data the model cannot see. The core aim of this paper is to propose a new class of neural networks able to model epistemic learning in a random set/belief function framework. We call this approach \emph{epistemic deep learning}, and argue that a deep neural network producing an outcome for (selected) subsets of the target space is an intrinsically more faithful representation of the epistemic uncertainty associated with the limited quantity and quality of the training data.

We first illustrate the principles of epistemic deep learning (Section \ref{sec:framework}).
A key step towards this is re-defining the ground truth of observations to represent the set of all subsets of the relevant target space (i.e., its \textit{power set}). 

\subsection{Formal framework} \label{sec:framework}

\subsubsection{Epistemic artificial intelligence}

\begin{figure}[ht!]
\centering
\includegraphics[width = \textwidth]{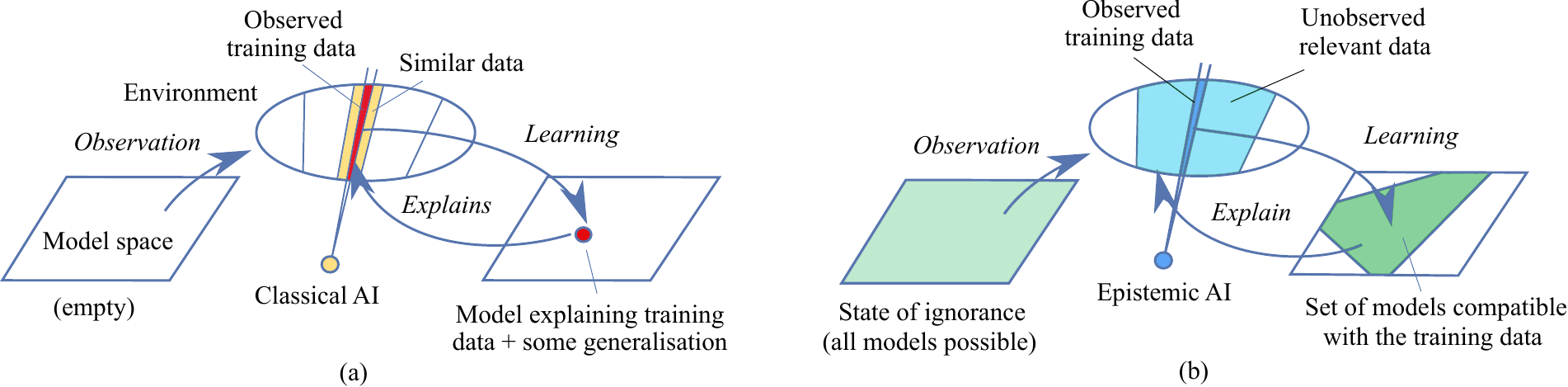}
\caption{Epistemic AI’s notion of learning (b), as opposed to that of traditional machine learning/artificial intelligence (a).} \label{fig:epi}
\end{figure}

The principle of \emph{epistemic artificial intelligence} is illustrated in Figure \ref{fig:epi}.
While traditional machine learning learns from the (limited) available evidence a model is able to describe with limited power of generalisation (a), epistemic AI (b) starts by assuming that the task at hand is (almost) completely unknown, because of the sheer imbalance between what we know and what we do not know. Our ignorance is then only tempered in the light of the (limited) available evidence to avoid forgetting how much we ignore about the problem or, in fact, that we even ignore how much we ignore. This principle translates into seeking to learn \emph{sets of hypotheses} compatible with the (scarce) data available, rather than individual models. A set of models may provide, given new data, a robust set of predictions among which the most cautious one can be adopted.

Mathematically, this can be done by modelling epistemic uncertainty via a suitable uncertainty measure on the target space $\Theta$ for the task at hand (e.g., the list of classes $\mathcal{C}$ for classification or a subset of $\mathbb{R}^n$ for regression problems).

\subsubsection{Target-level vs parameter-level representation}

While the rationale of epistemic deep learning (and epistemic AI more in general) is to model epistemic uncertainty using a suitable uncertainty measure, in practice this can happen at two distinct levels:
\begin{itemize}
    \item 
    A \emph{target level}, where the neural network is designed to output an uncertainty measure defined on the target space at hand, but its parameters (weights) are deterministic.
    \item
    A \emph{parameter level}, in which the network is modelled by an uncertainty measure on the parameter space itself (i.e., its set of weights).
\end{itemize}

In what follows we will focus, in particular, on target-level representations (see Sec. \ref{sec:rs-cnns}). However, the full generality of the framework can only be expressed at parameter level, where the purpose of training is to identify an uncertainty measure on the space of weights.

\subsubsection{Training objectives}

The two modelling levels are reflected in their training objectives. At target level, a loss function needs to be defined to measure the difference between predictions and ground truth, both in the form of uncertainty measures on the target space. However, this loss is a function of a deterministic weight assignment. In other words, target-level epistemic learning generalises traditional deep networks which output point predictions over set-valued ones.

At parameter level, the objective is to recover an uncertainty measure defined on the set of weights, or more generally, the hypothesis space for that class of models. For a given weight configuration, we can assume that the output is of the same kind as a classical deep network: for instance, for classification, a set of scores that can be calibrated to a probability distribution over the target space.

As a consequence, parameter-level epistemic networks will have the same output layers as traditional networks, but different in the training process. Target-level networks will have different output layers designed to output a more general uncertainty measure than a classical distribution.

%% file: randomsetcnn.tex
\section{Random-set convolutional neural networks} \label{sec:rs-cnns}
Having discussed the principles of epistemic deep learning, we wish to propose a design for a \emph{random-set convolutional neural network} that can be trained to output scores for sets of outcomes, and encode the epistemic uncertainty associated with the prediction in the random set framework. Note that, in this initial paper, we only focus on target-level representations.

\subsection{Random sets and belief functions}

Let us denote by $\Omega$ and $\Theta$ the sets of outcomes of two different but related problems $Q_1$ and $Q_2$, respectively. Given a probability measure $P$ on $\Omega$, we want to derive a `degree of belief' $Bel(A)$ that $A\subset \Theta$ contains the correct response to $Q_2$. If we call $\Gamma(\omega)$ the subset of outcomes of $Q_2$ compatible with $\omega\in\Omega$, $\omega$ tells us that the answer to $Q_2$ is in $A$ whenever $\Gamma(\omega) \subset A$.
The \emph{degree of belief} $Bel(A)$ of an event $A\subset\Theta$ is then the total probability (in $\Omega$) of all the outcomes $\omega$ of $Q_1$ that satisfy the above condition \cite{Dempster67}:
\begin{equation} \label{eq:belvalue}
Bel(A) = P(\{ \omega | \Gamma(\omega) \subset A \}) = \sum_{\omega\in\Omega
: \Gamma(\omega)\subset A} P(\{\omega\}).
\end{equation}
The map $\Gamma : \Omega \rightarrow 2^{\Theta} = \{A \subseteq \Theta\}$ 
is called a \emph{multivalued mapping} from $\Omega$ to $\Theta$. Such a mapping, together with a probability measure $P$ on $\Omega$, induces a \emph{belief function} on $2^\Theta$.
In Dempster's original formulation, then, belief functions are objects induced by a source probability measure in a decision space for which we do not have a probability, as long as there exists a 1-many mapping between the two.

Nevertheless, belief functions can also be defined axiomatically on the domain of interest (\emph{frame of discernment}) $\Theta$, without making reference to multivalued mappings.
A \emph{basic probability assignment} (BPA) \cite{Shafer76} is a set function \cite{denneberg99interaction,dubois86logical} $m : 2^\Theta\rightarrow[0,1]$ s.t.$m(\emptyset)=0$ and
$\sum_{A\subset\Theta} m(A)=1$.
In Dempster's interpretation, the `mass' $m(A)$ assigned to $A$ is in fact the probability $P(\{\omega \in \Omega : \Gamma(\omega) = A\})$. Shafer and Smets \cite{smets91other}, amongst others, have supported a view of mass functions as independently defined on $\Theta$. Subsets of $\Theta$ whose mass values are non-zero are called \emph{focal elements} of $m$.

The \emph{belief function} (BF) associated with a BPA $m : 2^\Theta\rightarrow[0,1]$ is the set function $Bel : 2^\Theta\rightarrow[0,1]$ defined as:
\begin{equation} \label{eq:belief-function}
Bel(A) = \sum_{B\subseteq A} m(B). 
\end{equation}
The corresponding \emph{plausibility function} is 
$Pl(A) \doteq \sum_{B\cap A\neq \emptyset} m(B) \geq Bel(A)$. A further equivalent definition of belief functions can be provided in axiomatic terms \cite{Shafer76}.
Classical probability measures on $\Theta$ are a special case of belief functions (those assigning mass to singletons only), termed \emph{Bayesian belief functions}. 

Based on these principles and definitions, we propose the following architecture and training objectives for a novel class of models called \emph{random-set neural networks}.

\subsection{Architecture} \label{sec:architecture}

Consider a simple classification neural network. A classifier $e$ is a mapping from an input space $X$ to a categorical target space $Y=[N]$, where $N$ denotes the number of classes:
\[e: X \rightarrow [N]\]
In set-valued classification, on the other hand, $e$ is a mapping from $X$ to the set of all subsets of $[N]$, the powerset $\mathbb{P}(N)$. 
\[e: X \rightarrow 2^{[N]} \rightarrow \mathbb{P}(N)\]

\begin{figure}[htp]
 \includegraphics[width=1\textwidth,keepaspectratio]{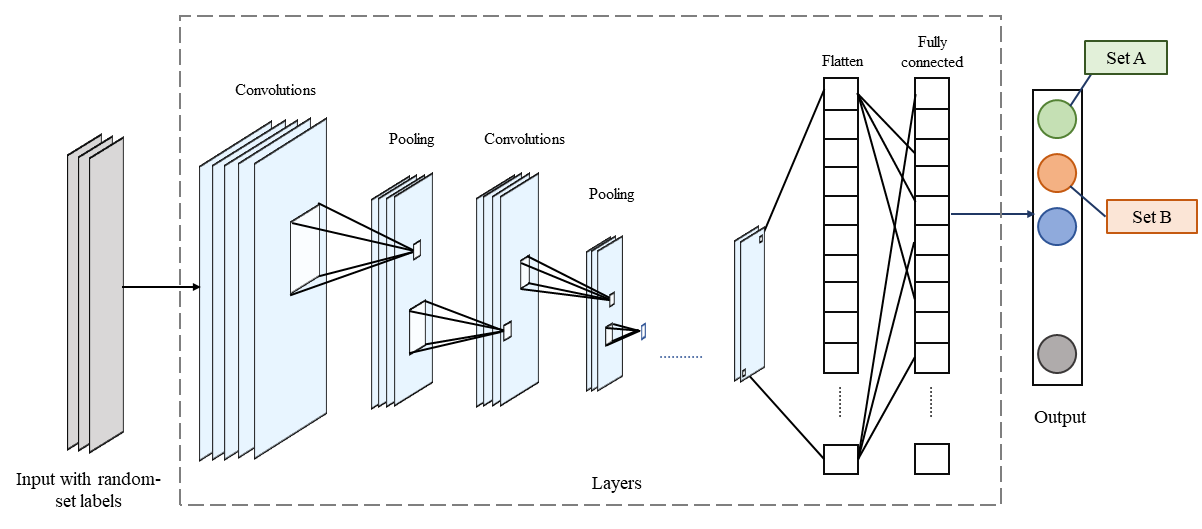}
 \caption{Random-set convolutional neural network}
\end{figure}

To train a network based on the above definition, we first modify the traditional ground truth vector \textbf{{$\hat y$}} for each data point $x$ to produce a set-valued output. This can be done in two ways:
\begin{itemize}
 \item By encoding the mass value $m(A)$ for each subset $A$ of the power set  $\mathbb{P}(N)$ of $N$ classes. 
 \[\hat y = \set{ m(A); A\in \mathbb{P}(N) } \]%
 \item By encoding the belief value $Bel(A)$ for each subset $A$ of the power set  $\mathbb{P}(N)$. 
 \[\hat y = \set{ Bel(A); A\in \mathbb{P}(N) } \]%
 \end{itemize}

For a list of N classes, a random-set CNN will contain, if naively implemented, $2^N$ classes with an output score for each subset of $\mathbb{P}(N)$. Using the mass value encoding, the ground truth vector is $GT=\{m_1,m_2,\dots,m_{2^N}\}$, where $m_i$ is the mass value for $i\in\mathbb{P}(N)$, whereas in traditional CNNs, $GT=\{1,2,\dots,N\}$. In the belief value encoding, the target vector becomes $GT=\{Bel_1,Bel_2,\dots,Bel_{2^N}\}$, where $Bel_i$ is the belief value for $i\in\mathbb{P}(N)$.

\subsection{Training} 
By re-defining the ground truth to represent set-valued observations, the model is able to learn set-structure and better estimate uncertainty. The mass value encoding produces similar results to a traditional CNN in terms of training and evaluation, as it simply looks like a padded version of the original target vector, with zeros added in correspondence of all non-singleton sets, allowing us to plug in as loss function the standard cross-entropy loss. The belief value encoding, instead, has a more complex structure since $Bel(A)$ for set $A$ is the sum of the masses of its subsets, but also the potential to “teach” the network the set structure of the problem.

Outputting scores for sets of target values is the basis for epistemic learning. The question then becomes what loss functions to use for training an epistemic neural network, by backpropagating gradients based on the different between predictions and ground truth.

%% file: loss.tex

\section{Epistemic loss functions} \label{sec:losses}

As a random-set network outputs scores for sets of outcomes, traditional loss functions such as cross-entropy cannot be used as an objective.



This provides two possible ways of generalising the cross-entropy loss to random-set networks: by employing suitable entropy measures for belief functions, or by generalising the concept of Kullback-Leibler (KL) divergence to them.

\subsection{Generalising the Kullback-Leibler divergence} \label{sec:kl}

The KL divergence of two probability distributions $P$ and $Q$ is defined as: 
\begin{equation}
    D_{KL}(P||Q)=\int_{x}P(x)\log\frac{P(x)}{Q(x)}\,dx.
\end{equation}

We wish to extend the above definition to belief functions defined on the target space, such as those outputted by a random-set network.

\subsubsection{KL divergence in the random set setting}

\begin{figure}[h!]
 \centering
 \includegraphics[width=1\textwidth,keepaspectratio]{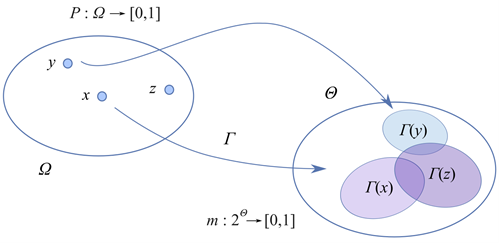}
 \caption{A multivalued mapping linking a probability on a source space $\Omega$ to a belief function on the decision space $\Theta$. \label{fig:multivalued}}
\end{figure}

The problem can be best posed in the random-set interpretation of belief functions, in which the latter are measures induced in the decision space of interest by a multi-valued mapping applied to a “source” domain in which source probabilities are available (see Figure \ref{fig:multivalued}) \cite{cuzzolin2020geometry}.

Let $\Gamma:\Omega\rightarrow2^\Theta$ be the multivalued (1-to-many) mapping from the (unknown) source space $\Omega$ to the set $2^\Theta$ of all subsets of the decision space $\Theta$ (set of classes). Let $m=\Gamma(P)$ and $\hat m=\Gamma(Q)$ be two mass functions on $\Theta$ encoding, respectively, the true class value for an input $x$ and the corresponding epistemic prediction generated by a network. The two are in fact images of two source probabilities in the unknown source space $\Omega$, which we call $P$ and $Q$.

\subsubsection{Approximating the KL divergence of the source probabilities}

The question is, if we plug in the mass function encoding, where $m(x)=P(x)$ is the true value and $\hat m(x)=Q(x)$ is the predicted value of the network, in the above standard definition of $D_{KL}$, do we obtain a good approximation of the KL divergence between the associated source probabilities $P$ and $Q$ in the source space $\Omega$, $D_{KL}(m||\hat m)\thickapprox D_{KL}(P||Q)?$ 

Since the KL divergence of mass functions, 
\begin{equation}
\sum_{A\subset\Theta}m(A)\log\frac{m(A)}{\hat m(A)} = \sum_{A\subset\Theta}\sum_{x:\Gamma(x)=A}P(x)\log\frac{\sum_{x:\Gamma(x)=A}P(x)}{\sum_{x:\Gamma(x)=A}Q(x)},
\end{equation}

whenever each subset $A\in\Theta$ is image of a unique point $x^*$ of $\Omega$, $D_{KL}(m||\hat m)=D_{KL}(P||Q)$. Thus, $D_{KL}(m||\hat m)$ is in general a reasonable approximation of the true KL divergence of the source probabilities. Unfortunately, the relationship is much more complicated with $D_{KL}(Bel||\hat{Bel})$ because of the combinatorial structure of belief values. 

We are currently working along these lines to provide a theoretical support for possible candidate extensions, and analytical solutions to optimisation problems involving generalised KL divergences as the objective function, under additional constraints.

\subsection{Generalised entropy measures} \label{sec:entropy}

The same question, in fact, applies to the generalisation of cross-entropy via entropy measures. The question of defining a suitable entropy measure for belief functions has been rather extensively studied in the literature \cite{nguyen}. A suitable candidate should reduce to the standard Shannon entropy when applied to probability measures. In addition, desirable properties an entropy measure should obey can be formulated \cite{JIROUSEK201849}.

In our work we have considered the following measures:

\subsubsection{Nguyen's entropy} \label{sec:nguyen}

First, we consider Nguyen’s generalisation of Shannon’s entropy \cite{Nguyen85entropy}  where probability values are replaced by mass values, 
\begin{equation} \label{eq:nguyen}
H_n[m] = -\sum_{A \in \mathcal{F}}m(A)\log⁡m(A),
\end{equation}

where $\mathcal{F}$ is the set of focal elements of $m$. Minimizing Nguyen's generalization as loss in random-set CNN produces satisfactory training loss metrics.

\subsubsection{Hohle’s entropy}
In Hohle’s measure of confusion \cite{Hohle1984}, predictive probabilities are replaced by belief values:

\begin{equation} 
H_h[m] = −\sum_{A\in F}m(A)\log{Bel(A)}
\end{equation}

\subsubsection{Yager’s entropy}
In Yager's entropy \cite{yager83entropy}, predictive probabilities are replaced by plausibility values:
\begin{equation}
H_y[m] = −\sum_{A\in F}m(A)\log Pl(A)
\end{equation}

\subsubsection{Klir and Ramer’s global uncertainty measure }
Later, Klir and Ramer \cite{klirramer} discussed some limitations of Hohle's measure of confusion \cite{Hohle1984}, and proposed a new composite measure called global uncertainty measure by combining Dubois and Prade's non-specificity \cite{doi:10.1080/03081078508934893} and a new measure of conflict called discord. 
\begin{equation} \label{eq:7}
H_k[m]=D[m]+H_d[m],
\end{equation}
where 
\begin{equation}
D(m)=-\sum_{A\in F}m(A)\log\left[⁡\frac{\sum_{B\in F}m(B)|A\cap B|}{|B|}\right] .
\end{equation}
The first component in Eq. \ref{eq:7} is designed to measure conflict, and the second component is designed to measure non-specificity.

\subsubsection{Desired properties of entropy for belief functions}
Jiroušek and Shenoy \cite{JIROUSEK201849} describes six desirable properties of entropy in the Dempster-Shafer(DS) theory, $H(m)$, where $m$ is the basic probability assignment:
\begin{itemize}
    \item 
    \emph{Consistency with DS theory semantics:} $P_{m_1\oplus m_2} = P_{m_1}\otimes P_{m_2}$, where $P_m$ is the probability mass function(PMF) transformed from $m$.
    \item
    \emph{Non-negativity:} $H(m)\geq 0$, equal if $x\in \Omega$, where $\Omega$ is the state space such that $m({x}) = 1$.
    \item
    \emph{Maximum entropy:} $H(m)\leq H(\iota)$, equal if $m=\iota$, where $\iota$ is the vacuous bpa. 
    \item
    \emph{Monotonicity:} If $|\Omega_X| < |\Omega_Y|$, then $H(\iota_X) < H(\iota_Y)$, where $X$ and $Y$ are random variables with state spaces $\Omega_X$ and $\Omega_Y$ 
    \item
    \emph{Probability consistency:} $H(m) = H_s(P)$, where $P$ is the PMF corresponding to $m$ and $H_s(P)$ is Shannon’s entropy of the PMF $P$. 
    \item
    \emph{Additivity:} $H(m_X\oplus m_Y) = H(m_X) + H(m_Y)$ for distinct BPAs $m_X$ and $m_Y$ for $X$ and $Y$ respectively.
\end{itemize}
Nguyen's entropy $H_n[m]$, Hohle's measure $H_h[m]$ and Yager's entropy $H_y[m]$ captures only the conflict portion of uncertainty. Therefore, they do not satisfy non-negativity, maximum entropy and monotonicity properties as $H_n(\iota)$, $H_h(\iota)$ and $H_y(\iota)$ are equal to zero. 

As $H_n$ follows the properties of Shannon's entropy, it satisfies the the consistency with DS theory semantics and probability consistency property. $H_n$ also satisfies the additivity property since the log of a product is the sum of the logs. Recall that for Bayesian BPA, $m({x}) = Bel_m({x})$ and $m({x}) = Pl_m({x})$. Therefore, $H_h$ and $H_y$ satisfies the consistency with DS theory semantics, probability consistency and the additivity property.

Klir and Ramer's $H_k[m]$ violates the maximum entropy property but satisfies the consistency with DS theory semantics, non-negativity, monotonicity, probability consistency, and additivity properties.

\subsection{Distance measures} \label{sec: distances}

A final approach consists of formulating a loss function based on a suitable distance function for belief functions, rather than attempting to generalise the notion of KL divergence. In this paper, we only consider one such approach:

\subsubsection{Jousselme's distance}

Jousselme's distance \cite{Jousselme200191} is based on the similarity among subsets and defines the distance between two BPAs $m_1$ and $m_2$ on the same frame of discernment (where \textbf{m} is represented as a vector) as:
\begin{equation}
d(m_1,m_2)=\sqrt{\frac{1}{2}(m_1-m_2)^T D(m_1-m_2)},\end{equation}
where
\[D(A,B)=\frac{|A\cap B|}{|A\cup B|}, A,B\in 2^\Theta\] 
is called the similarity measurement matrix. 

%% file: evaluation.tex
\section{Evaluation of epistemic predictions} \label{sec:evaluation}

As epistemic learning maps data points to sets of models, epistemic predictions cannot simply be evaluated using traditional performance measures, as these assume point-wise predictions, in turn computed from vectors of scores that can be calibrated to a probability distribution.

\subsection{Performance evaluation} \label{sec:evaluation-performance}

\subsubsection{Traditional deep learning} 

In classification problems, this is typically obtained by extracting the maximum-likelihood class from a vector $p$ of predicted probabilities of the various classes $c_i\in C$:

\[
\hat c = arg \max_i p(c_i) = arg \min_i d(p,p_i)
\]

where $p_i$ is a one-hot vector encoding class $c_i$ (equal to 1 for the true class and 0 otherwise), corresponding to a vertex of the simplex of probability distributions on $C$, and $d$ is a suitable distance function (e.g., an $L_p$ distance). Accuracy and average precision (AP) are then computed using the resulting point-wise class predictions. Devising a suitable evaluation strategy is essential to allow a fair comparison between traditional (e.g., Bayesian) and epistemic prediction methods.

\subsubsection{Bayesian deep learning}
Bayesian deep networks compute the predictive distribution by \emph{weighted model averaging}. The most common performance metrics in the evaluation of Bayesian neural networks include predictive performance (the ability of the model to provide accurate answers) and model calibration(the confidence about its prediction). Predictive performance is calculated similarly to traditional neural networks by computing various distance, entropy and log losses. Model calibration is computed as a calibration curve that compares between predicted and observed probabilities to detect whether a model is underconfident or overconfident. We believe that the current evaluation metrics for Bayesian neural networks do not test the true nature of epistemic predictions and thus is a field with scope for more research.

\subsection{Performance evaluation in epistemic deep learning} \label{sec:evaluation-epistemic}

\begin{figure}[h!]
 \centering
 \includegraphics[width=0.5\textwidth,keepaspectratio]{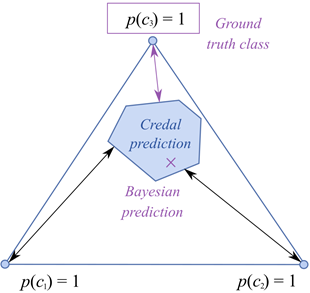}
 \caption{Proposed evaluation of epistemic predictions by minimising distances to the ground truth in the probability simplex. \label{fig:evaluation}}
\end{figure}

We have thus been studying the most appropriate ways to measure the performance of sets of models and of comparing it with that of traditional (single) models. We also look at theoretical guarantees of the dominance of epistemic predictions over Bayesian ones. To this end, we have been developing a general definition of performance based on measuring the distance (in the \textit{simplex} of possible probability distributions defined on the decision space or list of classes $C$) between the epistemic prediction (in the form of a credal set, either general or induced by a random set) and the ground truth (a one-hot probability corresponding to one of the vertices of the simplex): see Figure \ref{fig:evaluation}. 

Note that the size of the prediction should also be considered, for the model to learn the optimal trade-off between accuracy and precision of its predictions.

\subsubsection{Vertices of a random set prediction} \label{sec:vertices}

If the epistemic network outputs a belief function (finite random set) on the target space, before we measure any kind of distance between prediction and ground truth we first need to compute the vertices of the credal set \cite{levi80book,zaffalon-treebased,cuzzolin2010credal,antonucci10-credal} (i.e., a convex set of probabilities defined on the target space) \emph{consistent} with the predicted belief function.

Any probability distribution $P$ such that $P(A)\geq Bel(A)$ $\forall A \subset \Theta$, is said to be `consistent' with $Bel$ \cite{kyburg87bayesian}. Each belief function $Bel$ thus uniquely identifies a set of probabilities consistent with it,
\begin{equation} \label{eq:consistent}
\mathcal{P}[Bel] = \Big \{ P \in \mathcal{P} \Big | P(A) \geq Bel(A) \Big \},
\end{equation}
(where $\mathcal{P}$ is the set of all probabilities one can define on $\Theta$). 

Not all credal sets are consistent with a belief function. Credal sets associated with BFs have as vertices all the distributions $P^\pi$ induced by a permutation $\pi = \{ x_{\pi(1)}, \ldots, x_{\pi(|\Theta|)} \}$ of the singletons of $\Theta = \{ x_1, \ldots, x_n \}$ of the form \cite{Chateauneuf89,cuzzolin08-credal}
\begin{equation} \label{eq:prho}
P^\pi[Bel](x_{\pi(i)}) = \sum_{\substack{A \ni x_{\pi(i)}; \; A \not\ni x_{\pi(j)} \; \forall j<i}} m(A).
\end{equation}
Such an extremal probability (\ref{eq:prho}) assigns to a singleton element put in position $\pi(i)$ by the permutation $\pi$ the mass of all the focal elements containing it, but not containing any elements preceding it in the permutation order \cite{wallner2005}.

Eq. (\ref{eq:prho}) analytically derives the finite set of vertices which identify a random-set prediction. As a convex set can be handled using only its vertices, any distance between the ground truth and a random-set prediction will be a function of these vertices.

%% file: experiments.tex
\section{Experiments} \label{sec:experiments}
We have performed experiments for random-set convolutional neural networks and their mass-function encodings on four classification datasets and compared the results to traditional and Bayesian neural networks. We also discuss experiments on possible loss functions for random-set CNNs.
\subsection{Experiments on random-set CNNs}

\subsubsection{Binary classification}
\subsubsection*{Pima Indian Diabetes}
In our first set of evaluations on a binary classification problem, we trained on the \textit{Pima Indians Diabetes} dataset \cite{pima}, which consists of 8 predictor variables and one target variable (‘Outcome’) to predict whether a patient has diabetes or not (binary classification). The classes are represented as 0 and 1. We consider the power set of these classes which are set-valued classes of $\varnothing,{[0]},{[1]}$ and $[{0,1}]$. We then modify the ground truth from class representations (labels) of 0 and 1 to the mass values of the power set of classes, $m(\varnothing),m([0]),m([1]),m([0,1])$, and train the neural network using standard cross-entropy with the modified labels to output scores for each of the subsets of the power set. 

The random-set neural network with mass value as ground truth produced a test accuracy of $72.9\%$ when compared to the traditional neural network with an accuracy of $73.4\%$ and the Bayesian network with an accuracy of $67.7\%$. 

\subsubsection{Multi-class image classification}
\subsubsection*{MNIST}
In a second evaluation setting on multi-class image classification, we train the proposed random-set convolutional neural network for 4 out of 10 classes on the \textit{MNIST dataset}\cite{lecun-mnisthandwrittendigit-2010}. 

Our CNN architecture employs 32 filters of size 3×3 in the first convolutional layer, 64 filters of size 3×3 in the second and third convolutional layers, and 100 hidden units for the fully-connected (classification) layer. MNIST contains 60,000 training images and 10,000 testing images for 10 classes of images of digits from 0 to 9, corresponding to the class labels. In order to limit the computational issues, we choose 4 out of these 10 classes to train our network, thus forming a smaller dataset consisting of 24,754 training images and 4,157 testing images for classes 0 to 3, with a power set of cardinality $2^4=16$ (including the empty set). Similarly to what was mentioned above, we modify the ground truth to the mass values of each subset in the power set and use these as our new labels to train the network. 

We train the random-set CNN over 50 epochs to minimise the cross entropy loss between the true and predicted values. Table \ref{tab:table1} shows the test accuracies for MNIST dataset using traditional, Bayesian and random-set convolutional neural networks. Training the random-set CNN with cross-entropy as loss yields a test accuracy of $99.95\%$ when compared to the traditional CNN, which produced a test accuracy of $99.87\%$, and the Bayesian CNN with an accuracy of $98.28\%$ in the same experimental setup.

\subsubsection*{Intel Image Classification}

We also look at datasets less saturated than MNIST to evaluate our random-set CNN classifiers, in order to demonstrate its performance in more challenging settings. One of them is the Intel Image Classification dataset \cite{intelimage}, with 14,034 training, 3,000 testing and 7,000 prediction images over 6 classes. The random-set CNN architecture for this dataset has 32 filters of size 3×3 in the first convolutional layer, 64 filters of size 3×3 in the second and third convolutional layers, 126 filters of size 3×3 in the fourth and fifth layers and 512 hidden units for the fully-connected layer. Once again, we take into account 4 of the 6 classes. For a cross-entropy loss, the random-set CNN produces a test accuracy of $87.63\%$, the traditional CNN produces a test accuracy of $84.70\%$, and the Bayesian CNN produces an accuracy of $73.62\%$.

\subsubsection*{CIFAR10}
Another dataset of interest in the CIFAR-10 dataset \cite{cifar10} which consists of 60,000 32x32 color images divided across 10 classes, with 6000 images per class. These are split into 50,000 training images and 10,000 test images. Out of 10 classes, we consider 4 classes and modify the ground truth to represent the mass values of subsets of the power set of these classes. We use a similar architecture to that of MNIST to train this dataset and obtain a test accuracy of $85.90\%$ when compared to the traditional CNN, which produced a test accuracy of $84.95\%$, and the Bayesian CNN with an accuracy of $69.74\%$ in the same experimental setup.

\begin{table}
  \centering
  \begin{tabular}{c c c c}
  \hline
  \textbf{Datasets} & \textbf{Traditional} & \textbf{Bayesian} & \textbf{Epistemic}\\ 
  \hline
      \textit{Pima Indian Diabetes} & 73.43 & 67.70 & 72.91\\ 
      \textit{MNIST} & 99.87 & 98.29 & \textbf{99.95}\\ 
      \textit{Intel Image Classification} & 84.70 & 73.62 & \textbf{87.63}\\ 
      \textit{CIFAR10} & 84.95 & 69.74 & \textbf{85.90}\\ 
    \hline
    \end{tabular}
    \caption{Comparison of test accuracies(\%) in traditional, Bayesian and Epistemic convolutional neural networks (CNNs)}
    \label{tab:table1}
\end{table}

\subsection{Experiments on generalization of losses for random-set CNNs}

We also conducted experiments to find suitable loss functions for the proposed random-set CNNs by employing entropy measures. These experiments were carried out on the MNIST dataset for random-set CNNs.

\subsubsection{Kullback-Leibler divergence}

Training the random-set CNN with the Kullback-Leibler(KL) divergence divergence of the true and predicted masses as loss produces a test accuracy of $99.92\%$ over a traditional CNN with accuracy of $99.87\%$. This result is due to the mass function approximation of KL divergence discussed in Sec. \ref{sec:kl}

\subsubsection{Entropy measures}
Among the different entropy measures discussed in Sec. \ref{sec:entropy}, we have obtained comparable results for \emph{Nyugen's entropy} (see Sec. \ref{sec:nguyen}), which is a generalization of Shannon's entropy. A random-set CNN with Nguyen's entropy as loss yields a test accuracy of $99.85\%$ and a test loss of $0.0017\%$ on the MNIST dataset. The mass function encoding of random-set CNNs is similar to the original target vector with zeros added in correspondence of all non-singleton sets. This is one of the reasons in obtaining good accuracy results when compared with traditional and Bayesian neural networks.
In on-going work, we are performing expriments on the other entropy measures mentioned in Sec. \ref{sec:entropy}, namely Hohle's measure, Yager's entropy and Jousselme's distance(see Sec. \ref{sec: distances}).

\subsection{Results}

As argued above, our conclusion is that a mass function encoding of the output of a random-set CNN is similar to a traditional output in terms of training and validation. More sophisticated techniques, such as the belief function encoding (discussed in \ref{sec:architecture}) can be used to teach the network the power set structure of the target-level representation. This approach is being explored in on-going work. We are also looking at methods to effectively select the most significant classes within the power set for training. Using a power set structure for set-valued observations increases the complexity of the network exponentially. Since our current focus is not on the complexity of the representation, we only use a few of the classes to carry out the experiments. Despite these, random-set CNNs have produced better test accuracies and test losses than traditional and Bayesian neural networks in most of the datasets discussed in Table \ref{tab:table1}.

%% file: conclusion.tex
\section{Conclusions} 
\label{sec:conclusions}

In this paper, we develop a novel approach to estimating uncertainty using the random set/belief function approach called \emph{epistemic deep learning}. We propose a random-set convolutional neural network for classification using this approach and demonstrate its performance by comparing with traditional and Bayesian neural networks. We discuss possible loss functions for use in these networks and performance evaluation metrics for epistemic predictions. Our method produces better test accuracies for classification tasks when compared to current state-of-the-art neural networks for uncertainty estimation and is comparable to traditional networks.

In this work, we only discuss target-level representation of random-set CNN. We are currently looking at the parameter-level representation of random-set neural networks in on-going work and extending this approach to detection tasks. Detection networks have additional heads which take care of regressing bounding box coordinates in addition to labels: the research problem is then based on how to formulate suitable localisation losses in keeping with the epistemic uncertainty principle, e.g. by outputting a random set over the search space of possible bounding boxes.

Current classification models and benchmark datasets are designed for classical, single-class ground truth representations per data point. The challenge is then to define and train the neural network to learn random sets and the subset structure of the problem, the $2^N$ ground truth representation in epistemic learning. As discussed under Sec. \ref{sec:evaluation}, the performance and evaluation metrics for random set neural networks will also forcibly differ from that of traditional neural networks. Hence, it is important to derive a generalized evaluation framework to measure the performance of traditional, Bayesian and Epistemic predictions.